\documentclass[11pt]{article}
\usepackage{acl}
\usepackage{times}
\usepackage{latexsym}
\usepackage{graphicx}
\usepackage{booktabs}
\usepackage{amsmath}
\usepackage{enumitem}

\newcommand{\system}{\textsc{InvestMate}}

\title{High-Stakes Personalization: Rethinking LLM Customization for Individual Investor Decision-Making}

\author{Yash Ganpat Sawant \\
  AI Researcher \\
  \texttt{sawantyash13@gmail.com}}

\begin{document}
\maketitle

\begin{abstract}
Personalized LLM systems have advanced rapidly, yet most operate in domains where user preferences are stable and ground truth is either absent or subjective. We argue that individual investor decision-making presents a uniquely challenging domain for LLM personalization---one that exposes fundamental limitations in current customization paradigms. Drawing on \system{}, a system we built and deployed for AI-augmented portfolio management, we identify four axes along which individual investing exposes fundamental limitations in standard LLM customization: (1)~\textit{behavioral memory complexity}, where investor patterns are temporally evolving, self-contradictory, and financially consequential; (2)~\textit{thesis consistency under drift}, where maintaining coherent investment rationale over weeks or months strains stateless and session-bounded architectures; (3)~\textit{style--signal tension}, where the system must simultaneously respect personal investment philosophy and surface objective evidence that may contradict it; and (4)~\textit{alignment without ground truth}, where personalization quality cannot be evaluated against a fixed label set because outcomes are stochastic and delayed. We describe the architectural responses that emerged from building \system{} and propose open research directions for personalized NLP in high-stakes, temporally extended decision domains.
\end{abstract}

\section{Introduction}

The personalization of large language models has become a growing area of research, with systems that adapt to user writing style \cite{salemi2024lamp}, retrieve user-specific context for generation \cite{mysore2024pearl}, and maintain preference profiles across sessions \cite{zhang2025personalization}. Recent surveys frame personalized alignment as a ``missing piece'' for real-world LLM deployment \cite{guan2025personalized}, and benchmarks like PersonalLLM \cite{personalllm2025} have begun to formalize the evaluation of preference-conditioned generation.

These advances share a common setting: domains where the cost of misalignment is low, preferences shift slowly, and evaluation is largely subjective. Writing assistance, content recommendation, and conversational companionship---the dominant application areas---are forgiving environments for personalization research.

Finance, and specifically individual investor decision-making, is not. A personalized investment assistant must track evolving conviction about specific holdings, recall the reasoning behind decisions made weeks earlier, reconcile an investor's behavioral tendencies with objective market signals, and do all of this in a domain where outcomes are stochastic, delayed, and high-stakes. This combination of properties is not merely harder---it is \textit{structurally different} from what current personalization methods are designed for.

This position paper draws on our experience building \system{}, a system for AI-augmented individual portfolio management. It integrates living thesis management, daily conviction scoring, behavioral memory extraction, and drift detection into a thesis-centric architecture. It emerged not from a top-down design but from a practitioner's repeated encounters with the inadequacy of generic LLM interactions for investment decision support: losing context between sessions, receiving advice untethered from one's actual investment rationale, and watching models latch onto recency rather than the thesis that justified a position weeks ago.

\section{Background: What Makes Finance Different}

Most LLM personalization research assumes a relatively benign setting: users have stable preferences, the system's job is to match them, and evaluation reduces to preference satisfaction \cite{salemi2024lamp, personalllm2025}. Personalized retrieval \cite{mysore2024pearl}, user-conditioned generation \cite{chen2025survey}, and memory-augmented dialogue \cite{xu2022goldfish, zhong2024memorybank} all operate within this frame.

Meanwhile, financial NLP has focused on domain-specific language modeling \cite{wu2023bloomberggpt, yang2023fingpt} and, more recently, LLM-based agents for trading and portfolio construction \cite{li2025llmagents}. These efforts treat finance as a \textit{domain adaptation} problem. Personalization, when addressed at all, reduces to risk-profile elicitation \cite{sanzcruzado2025financial}.

Individual investing breaks both frames simultaneously:

\paragraph{Decisions are consequential and irreversible.} Unlike a misranked search result or an off-tone email draft, a poorly timed trade results in direct financial loss. The tolerance for personalization error is orders of magnitude lower.

\paragraph{Preferences are dynamic and self-contradictory.} An investor may state a rule (``never average down into a falling knife'') while exhibiting a pattern of doing exactly that during high-volatility periods. The system must represent both the stated rule and the revealed behavior---the tension between them is itself informative \cite{kahneman1979prospect, thaler2015misbehaving}.

\paragraph{Ground truth is delayed and stochastic.} A recommendation made today may take weeks to validate, confounded by market noise. There is no immediate label to train or evaluate against.

\paragraph{Temporal coherence is load-bearing.} An investment thesis stated six weeks ago must anchor today's evaluation. Stateless or session-bounded systems lose this thread entirely, defaulting to whatever narrative is most coherent at generation time.

\section{\system{}: A Motivating Architecture}

\system{} evolved from a simple portfolio tracker to a multi-layered personalization system. We describe its components not as contributions in themselves but as a lens through which the four challenge axes become concrete.

\subsection{Living Thesis Architecture}

The core abstraction is the \textit{living thesis}---a structured, per-holding hypothesis capturing not just what an investor owns but \textit{why}: a conviction statement, validation triggers, break conditions, macro dependencies, and upcoming catalysts. All downstream generation---reports, alerts, behavioral analysis---is \textit{thesis-aware}, referencing the investor's stated rationale rather than price data alone.

\subsection{Conviction Tracking and Behavioral Feedback}

Each thesis is evaluated daily against market signals through a focused LLM call returning a structured assessment: \texttt{CONFIRMED}~(+1), \texttt{UNCHANGED}~(0), \texttt{WEAKENED}~($-$1), or \texttt{BROKEN}~($-$2). Fixed deltas accumulate into a conviction trajectory---a time series of belief strength tied to specific evidence, replacing earlier unconstrained scores that drifted unpredictably.

This design replaced an earlier approach using unconstrained LLM-generated conviction scores, which drifted unpredictably and proved uninterpretable over time. The model would generate scores influenced by whatever market context felt most salient at generation time rather than by the actual thesis content. Fixed deltas with mandatory evidence grounding resolved this---each conviction change is auditable, anchored to a specific market event, and interpretable in the context of the original thesis.

Two mechanisms close the behavioral feedback loop. \textbf{Drift detection} flags when actions contradict stated theses---holding past break conditions, selling before validation triggers, or sizing inconsistently with conviction. \textbf{Pattern matching} mines journal entries, thesis evaluations, and closed positions to identify recurring behavioral tendencies, and grades closed positions (A--F) on \textit{process quality} rather than P\&L.

\subsection{Behavioral Memory}

A persistent behavioral profile is extracted from investor interactions via heuristic filtering for decision-relevant signals (e.g., ``my rule is...'', ``I always...''), with matched turns processed by an LLM into structured insights across five categories: \textit{preferences}, \textit{beliefs}, \textit{patterns}, \textit{rules}, and \textit{risk tolerance}. These insights are injected into all downstream contexts. As a concrete example: the system detected a pattern of buying into sharp price declines during high-volatility periods and began surfacing it alongside VIX readings---during a period of geopolitical escalation, this behavioral memory tempered what would otherwise have been an impulsive averaging-down decision.

\section{Four Axes Where Personalization Breaks}

\subsection{Axis 1: Behavioral Memory Complexity}

Standard user modeling treats profiles as relatively stable attribute sets \cite{salemi2024lamp, zhang2025personalization}. In investing, the behavioral profile is \textit{layered}, \textit{contradictory}, and \textit{consequential}:

\begin{itemize}[nosep,leftmargin=*]
    \item A \textit{stated rule} (``never chase falling knives'') may directly contradict a \textit{revealed pattern} (consistently buying during sharp declines). The system must represent both without resolving the contradiction---the tension is the signal.
    \item Memory insights decay at different rates: risk tolerance is durable over months; a belief about a specific earnings cycle is ephemeral over days.
    \item Extracting decision-relevant signals requires domain-aware heuristics. The phrase ``I'm going to hold'' carries fundamentally different weight in an investment context than in general dialogue.
\end{itemize}

In practice, naive memory approaches---storing all interactions or compressing history---produced profiles either too noisy to influence generation or too compressed to preserve the contradictions that matter. The five-category taxonomy emerged from iterating on what actually changed downstream outputs, suggesting that effective behavioral memory in high-stakes domains may require domain-specific ontologies rather than general-purpose architectures \cite{zhong2024memorybank}.

\subsection{Axis 2: Thesis Consistency Under Drift}

Long-term memory in dialogue has been studied extensively \cite{xu2022goldfish, zhong2024memorybank}, but the temporal consistency problem in investing is qualitatively different. It is not about \textit{recalling} past information---it is about \textit{evaluating present evidence against a past commitment} while distinguishing legitimate thesis evolution from unconscious drift.

Consider: an investor establishes a thesis on a semiconductor stock based on expected AI infrastructure spending. Six weeks later, earnings disappoint. A stateless LLM will generate a bearish assessment anchored to recent evidence. A thesis-aware system must ask: does this earnings miss \textit{break} the original thesis (AI capex cycle has peaked), or is it noise within an intact thesis (one quarter of digestion in a multi-year buildout)?

This distinction requires temporal grounding that current personalization architectures do not support. RAG retrieves relevant documents but does not enforce evaluative consistency. Summarization compresses away the specific validation and break conditions that make a thesis actionable. Structured thesis formats with explicit fields for triggers and break conditions are necessary precisely because free-text descriptions allow models to reinterpret the thesis toward whatever narrative is locally coherent.

It is worth distinguishing this from the factual consistency problem NLP has studied extensively. Hallucination detection asks whether a model's output is consistent with source documents. Thesis consistency under drift asks whether a model's \textit{evaluation} is consistent with a prior human commitment---a harder form of consistency with no ground-truth document to check against, and one where the ``correct'' answer depends on the investor's original intent rather than any external fact.

\subsection{Axis 3: Style--Signal Tension}

In most personalization settings, aligning with user preferences \textit{is} the objective \cite{guan2025personalized}. In investing, a system that consistently confirms the investor's priors is \textit{actively harmful}---this is the confirmation bias that behavioral finance has studied for decades \cite{kahneman1979prospect}.

The system must simultaneously respect the investor's framework (risk tolerance, analytical approach, conviction formation process) and challenge it when market evidence contradicts their thesis or actions are inconsistent with stated rules. This is not a simple ``sometimes agree'' heuristic---it requires understanding the framework well enough to know when deviation is warranted.

Sanz-Cruzado et al.~\cite{sanzcruzado2025financial} found that users preferred LLM advisors with extroverted personas even when those agents gave \textit{worse} advice---suggesting satisfaction and quality can be inversely correlated. Our system manages this tension architecturally: generation is anchored portfolio-first but explicitly injects drift warnings and behavioral pattern observations as counterpoints, encoding the dual mandate directly into the generation protocol.

\subsection{Axis 4: Alignment Without Ground Truth}

Personalization evaluation typically relies on preference ratings, A/B tests, or task completion metrics \cite{salemi2024lamp, personalllm2025}. In investing: outcomes materialize over months; good process regularly produces bad outcomes; the counterfactual is unobservable; and LLMs themselves exhibit cognitive biases that interact with the investor's own \cite{xie2024cogbias}.

Our approach to thesis grading addresses this by evaluating closed positions on \textit{process quality}: was the thesis directionally correct? Was timing appropriate? Was sizing consistent with conviction? A position that lost money but followed a sound thesis receives a higher grade than a profitable impulsive trade. This process-over-outcome framing is grounded in behavioral economics \cite{thaler2015misbehaving}, but formalizing it as a personalization evaluation metric remains an open problem.

The broader implication extends beyond finance: any domain where outcomes are stochastic and delayed---healthcare, educational guidance, career coaching---faces the same evaluation gap. Finance makes it impossible to ignore because P\&L is precise enough to \textit{seem} evaluable while being too noisy to serve as ground truth for personalization quality.

\section{Open Research Directions}

Our experience building such a system suggests several directions where the NLP community could contribute:

\paragraph{Temporal memory for evolving beliefs.} Current memory approaches \cite{zhong2024memorybank, xu2022goldfish} are designed for recall, not for maintaining structured commitments evaluated against evolving evidence. Memory architectures that natively represent belief trajectories, contradictions, and multi-rate decay would serve any domain with extended decision horizons.

\paragraph{Personalization under adversarial feedback.} Markets provide constant feedback that is noisy, delayed, and adversarial. Personalization methods that learn from such feedback without overfitting to outcome noise connect to RLHF under distribution shift---but with the added constraint that the human in the loop has biases the system must sometimes correct.

\paragraph{Process-quality evaluation frameworks.} We lack evaluation methodologies for personalized systems where ground truth is stochastic. Formalizing process-quality metrics---coherence, evidential grounding, and consistency with the user's stated framework---is an open problem that would benefit personalization research well beyond finance.

\paragraph{Controlled autonomy and constructive disagreement.} When should a personalized system defer to the user, and when should it push back? The style--signal tension raises fundamental questions about AI autonomy that, in financial contexts, carry direct economic consequences. Standard RLHF optimizes for user approval---exactly the wrong objective when user satisfaction and advice quality are inversely correlated \cite{sanzcruzado2025financial}. Developing principled mechanisms for constructive disagreement that preserve user trust while surfacing inconvenient evidence is an open problem connecting personalization research to the broader alignment literature.

\section{Conclusion}

Individual investor decision-making systematically stress-tests the assumptions underlying current LLM personalization research. Through building and deploying \system{}, we identified four axes---behavioral memory complexity, thesis consistency under drift, style--signal tension, and alignment without ground truth---along which generic customization breaks down. These challenges are not unique to finance; they arise wherever decisions are consequential, temporally extended, and evaluated against stochastic outcomes. We advocate for personalized financial AI that prioritizes transparency and auditability, framing outputs as decision support rather than directives. More broadly, we believe treating such domains as first-class personalization problems will drive more robust and general personalization architectures for the field.

\bibliography{references}

\end{document}